\documentclass[a4paper,10pt,twoside]{article}

\usepackage{clin}        
\usepackage{harvard}     

\usepackage{graphicx}
\usepackage{multirow}
\usepackage[dvipsnames]{xcolor}
\usepackage{booktabs}
\usepackage{caption}
\usepackage{subcaption}

\newcommand{\etc}{etc.\ }
\newcommand{\eg}{e.g., }

\newcommand{\ie}{i.e., }
\newcommand{\vs}{vs.\ }

\newcommand{\figref}[1]{Fig.~\ref{#1}}    
\newcommand{\Figref}[1]{Figure~\ref{#1}}  
\newcommand{\tabref}[1]{Table~\ref{#1}}
\newcommand{\Tabref}[1]{Table~\ref{#1}}
\newcommand{\secref}[1]{Section~\ref{#1}}
\newcommand{\equref}[1]{Eq.~(\ref{#1})}

\newcommand{\appref}[1]{Appendix~\ref{#1}} 

\newcommand{\Tag}[1]{\textless {#1}\textgreater} 
\usepackage[inline,shortlabels]{enumitem}
\usepackage[hidelinks]{hyperref}

\usepackage[colorinlistoftodos,prependcaption,textsize=normalsize]{todonotes}

\newcommand{\revchris}[1]{\textcolor{BlueViolet}{#1}}
\renewcommand{\revchris}[1]{\textcolor[rgb]{0,0.0,0.0}{#1}}
\newcommand{\reviewsemere}[1]{\textcolor{ForestGreen}{#1}}
\renewcommand{\reviewsemere}[1]{\textcolor[rgb]{0,0.0,0.0}{#1}}

\begin{document}

\title{Personality Style Recognition via Machine Learning: Identifying Anaclitic and Introjective Personality Styles from Patients' Speech}


\author{Semere Kiros Bitew$^*$ \email{semerekiros.bitew@ugent.be}\\
{\normalsize \bf Vincent Schelstraete}$^{***}$ \email{vincent@techwolf.be}\\
{\normalsize \bf Klim Zaporojets}$^*$ \email{klim.zaporojets@ugent.be}\\
{\normalsize \bf Kimberly Van Nieuwenhove}$^{**}$ \email{kimberly.vannieuwenhove@ugent.be}\\
{\normalsize \bf  Reitske Meganck}$^{**}$ \email{reitske.meganck@ugent.be}\\
{\normalsize \bf Chris Develder}$^{*}$ \email{chris.develder@ugent.be}
\AND \addr{$^{*}$IDLab, Department of Information Technology, Ghent University – IMEC, Ghent, Belgium}
\AND \addr{$^{**}$Department of Psycho-analysis and Clinical Consulting, Ghent University, Ghent, Belgium}
\AND \addr{$^{***}$TechWolf, Ghent, Belgium}}


\maketitle\thispagestyle{empty} 


\begin{abstract}
In disentangling the heterogeneity observed in psychopathology, personality of the patients is considered crucial. \revchris{While it has been demonstrated that personality traits are reflected in the language used by a patient, we hypothesize that this enables automatic inference of the personality type directly from speech utterances, potentially more accurately than through a traditional questionnaire-based approach explicitly designed for personality classification.} To validate this hypothesis, we adopt natural language processing (NLP) and standard machine learning tools for classification. We test this on a dataset of recorded clinical diagnostic interviews (CDI) on a sample of 79 patients diagnosed with major depressive disorder (MDD) -- a condition for which differentiated \revchris{treatment} based on personality styles has been advocated -- and classified into anaclitic and introjective personality styles. We start by analyzing the interviews to see which linguistic features are associated with each style, in order to gain a better understanding of the styles. Then, we develop automatic classifiers based on  \begin{enumerate*}[(a), afterlabel=\hspace{-1pt}]
    \item ~standardized questionnaire responses;
    \item~basic text features, \ie TF-IDF scores of words and word sequences; 
    \item ~more advanced text features, using LIWC (linguistic inquiry and word count) and context-aware features using BERT (bidirectional encoder representations from transformers);
    \item ~audio features.
\end{enumerate*} We find that automated classification with language-derived features (\ie based on LIWC) significantly outperforms questionnaire-based classification models. Furthermore, the best performance is achieved by combining LIWC with the questionnaire features. This suggests that more work should be put into developing linguistically based automated techniques for characterizing personality, however questionnaires still to some extent 
complement such methods.\footnote{Code will be available in \url{https://github.com/semerekiros/personapredict} - upon acceptance. } 
\end{abstract}

\section{Introduction}

Personality is considered a crucial variable in explaining the heterogeneity observed in psychopathology~\cite{blatt2004experiences}. 
It has been demonstrated that the large group of people meeting the commonly used criteria for depression (\eg DSM or ICD criteria) -- which affects over 280 million people worldwide \reviewsemere{(according to the WHO's 2021 report)},\footnote{\url{https://www.who.int/news-room/fact-sheets/detail/depression}} -- is not homogeneous.
On the contrary, the heterogeneity of people covered by the general label of depression is widely recognized \cite{goldberg2011heterogeneity}.

From multiple perspectives, a two-polarity model has been put forward to distinguish between depressed patient groups on the basis of two personality dimensions, namely relatedness and self-definition~\cite{blatt1974levels,beck1983cognitive,blatt2004experiences}. 
Drawing on Freud`s \cite{freud1953three} 
early distinction between hysteria (libidinal, oral fixation) and obsessional neurosis (aggressive, anal fixation), 
Blatt (1974, 2004)
distinguishes between an anaclitic and introjective personality style, respectively.
The \emph{anaclitic} personality style is focused on interpersonal relationships and dependency, whereas the \emph{introjective} personality style is focused on self-definition, autonomy, and interpersonal distance. 
Beck (1983)
 proposed similar interpersonal characteristics, sociotropy and autonomy, to discern different personality styles in depression.
 Sociotropy indicates an intense need for intimate relationships, whereas autonomy is associated with a desire for independence and achievement.
 As such, sociotropy and autonomy correspond with the anaclitic and introjective personality dimensions, respectively.
 Similar to Blatt (1974), Beck (1983) 
assumes that individuals are more prone to react to certain stressors, dependent on their personality style.
Anaclitic/sociotropic individuals are vulnerable for interpersonal stressors (\eg breakups), whereas introjective/autonomous individuals are more sensitive to stressors on the level of achievements and self-definition (\eg job loss).
\revchris{Consequently,} a predominance of either personality dimension leads to the development of distinct symptoms when confronted with these specific stressors\revchris{:} 
different phenotypical manifestations of depression have been described \revchris{from this perspective}.
The symptom-specificity hypothesis \cite{blatt2004experiences} 
assumes that anaclitic depression, typified by an overdetermination of dependency and a preoccupation with interpersonal consolidation, is associated with somatic symptoms and phobias.
An introjective depression on the other hand, is dominantly focused on self-definition and interpersonal distance and involves symptoms on a more cognitive level, exemplified by self-criticism, perfectionism and pathological doubt. 
\revchris{Noting these different manifestations,} it has been argued that \revchris{also} treatment should be tailored according to patients' personality style\revchris{s}, with anaclitic patients benefiting more from structured, supportive therapy approaches, whilst patients with an introjective personality style would profit more from insight-focused treatments \cite{werbart2017changes}. 
\revchris{Thus, a first fundamental step supporting this would be  effective determination of a patient's dominant personality style.}

Strikingly, notwithstanding their solid theoretical underpinnings, after almost half a century, personality style assessment remains a challenge, which is embodied by research failing to yield consistent support for the symptom-specificity hypothesis \cite{coyne2004promissory,desmet2007histerical} 
and outcome studies, as of yet, not having convincingly demonstrated the benefits of matching personality styles to different treatment strategies \cite{meganck2017ghent,werbart2018matching}. 
At the heart of this predicament probably lie the methodological difficulties associated with the assessment of personality styles \cite{desmet2007histerical}. 
The determination of personality styles typically includes the administration of self-report questionnaires, such as the Depressive Experience Questionnaire (DEQ) \cite{blatt1976experiences} 
and the Personal Style Inventory (PSI) \cite{robins1991sociotropy}. 
However, research literature is inconsistent as to whether these questionnaires are apt (valid) to distinguish between the clinical manifestations (symptoms, interpersonal functioning) associated with an anaclitic and introjective personality style \cite{coyne2004promissory,desmet2007histerical}. 
To make further progress, there is a need to explore alternative approaches to detect personality styles with greater precision.
More recent approaches such as prototype matching \cite{werbart2014changes} 
are promising, since they combine scientific rigor with clinical expertise.
Yet, they are time consuming because they rely on the expertise of several well-trained human raters. 
Natural Language Processing (NLP) and Machine Learning (ML) techniques, originating from the field of computer science, might alternatively provide ways to assess psychological features and phenotypes based on complex data without the interference of human raters \cite{dwyer2018machine}. 
NLP focuses on automated processing of human language, learning and applying the underlying linguistics and semantics in computerized systems \cite{joseph2016natural}. 
Today, NLP applications heavily rely on machine learning techniques.
Machine learning (ML) is a subset of artificial intelligence that ``automatically determines (\ie learns) methods and parameters to reach an optimal solution to a problem rather than being programmed by a human a priori to deliver a fixed solution'' \cite{dwyer2018machine}. 
In other words, ML offers techniques to identify patterns, phenotypes or subgroups in a dataset (\eg text, audio, self-report questionnaires, biological data) through learning from the data with minimal prior assumptions.
Although ML techniques are mostly known for their commercial (\eg personalized advertisements via internet searches) and medical applications (\eg algorithms to detect cancer), there is over two decades worth of research in the psychological field, concerning the diagnosis, prognosis and treatment of diverse mental conditions \cite{dwyer2018machine}, 
such as depression \cite{fu2019addressing} and suicide contemplation \cite{bitew2019predicting}.
Unfortunately, the use of mostly complex biological data (\eg neuroimaging and MRI data) complicates the translation of ML solutions to practical and clinical implementations \cite{dwyer2018machine}. 
In the context of psychotherapy research, however, what patients say about themselves and their symptoms is an essential, if not the most important source of information, which makes patients' narratives a particularly suitable source of data for NLP and ML.
For the detection of depression in general, there are several studies using NLP on textual data \cite{al2018detecting,coppersmith2015clpsych,ozkanca2018multi,tausczik2010psychological}. 
As an example, the Linguistic Inquiry and Word Count (LIWC), a text analysis program that counts the frequency of words and types of grammar typically used in certain contexts, 
Tausczik and Pennebaker (2010),
found that depressed people use more negative words and terms related to death on social media, while also using the first person more frequently than healthy controls.
Next to the detection of depression using textual information from social media \cite{coppersmith2015clpsych,tausczik2010psychological}, 
depression has been successfully inferred using transcripts from screening interviews \cite{al2018detecting} 
and therapy sessions \cite{ozkanca2018multi}. 
However, as 
Dwyer et al. (2018)
note, it does not suffice to use ML to detect broad categories (\eg depression yes/no), as treatment for depression in general shows relatively low efficacy \cite{driessen2015efficacy}. 
Especially for the heterogeneous depressed population, it would be fruitful to use ML techniques to predict more specific features, such as personality style, which would allow more tailored clinical decision making.

Thus, several works were proposed that employ ML for automatic personality prediction using different personality frameworks (measures) such as Big Five \cite{digman1990personality}, MBTI \cite{myers1988myers}, and Catell's 16PF \cite{cattell2008sixteen}. The majority of research makes use of the Big Five personality measure for personality trait classification \cite{majumder2017deep,schwartz2013personality,mairesse2007using,mairesse2006automatic,stachl2020personality} followed by MBTI \cite{yang2021multi,gjurkovic2018reddit}.
The Big Five personality measure defines personality through the following five dimensions: extroversion, agreeableness, conscientiousness, neuroticism, and openness.
Similar to depression detection systems, automatic personality detection ML models use a wide range of data sources; these sources include digital footprints in social media platforms (\eg Reddit \cite{gjurkovic2018reddit}, Twitter \cite{kalghatgi2015neural}, Facebook \cite{hall2017say,golbeck2011predicting}), essays written in a controlled environment by volunteers \cite{tausczik2010psychological,pennebaker1999linguistic}, video and audio recordings of meetings \cite{carletta2005ami}.
The input modalities that these ML models use include text, audio, visual cues, and multimodal inputs.
Many personality detection ML models that use \emph{textual data} as input extract features using two approaches
\revchris{:} closed-vocabulary and open-vocabulary.
Closed-vocabulary approaches rely on 
\revchris{a priori determined word categories, each represented by a fixed set of words (where some may belong to multiple categories).}
For example, LIWC, which is the most common closed-vocabulary method for personality detection from text, categorizes the words in a text into various psychologically relevant clusters like `affective processes' (\eg `happy', `cried', `nervous') and `social processes' (\eg `pal', `buddy', `cousin', `woman').
The 
\revchris{frequency counts of words for}
each cluster 
\revchris{are} used by ML models to predict the personality of the text author \cite{hall2017say}. 
On the other hand, open-vocabulary methods rely on extracting a comprehensive collection of language features from textual input such as n-gram features, part of speech (POS) tags, usage of emoticons, length of posts of users from social media, and use of word embeddings (\eg Word2vec, GloVe) 
to build \revchris{ML models for} personality detection 
\cite{schwartz2013personality,plank2015personality}. 
\revchris{Apart from the actual words spoken, additional features can be extracted from the \emph{audio signal}, \eg}
loudness, spectrals, pitch, interruptions, acoustic features, etc.
For instance, as the first stage in their deception detection pipeline, \cite{levitan2016identifying} use prosodic, acoustic, and LIWC features to predict the personality traits of speakers engaging in a dialogue.
They analyzed how each LIWC element contributed 
\revchris{and} discovered\revchris{, among others,} that the dimensions of ``focus-future'' and ``drives'' are the most helpful features for determining ``extroversion'' personality traits, whereas ``time'' 
\revchris{and} ``work'' are significant for determining ``conscientiousness.''
They also show that incorporating the personality traits as an input improves their deception detection model.

\emph{Visual-based} ML models for personality prediction mainly extract features of the body, especially the face.
Facial features such the shape of nose 
\revchris{and} eyebrows, eye openness, and mouth have been used to infer personalities of users \cite{kamenskaya2008recognition}.
For example, Liu et al. (2016) 
trained an ML model to predict personality of people by analyzing their Twitter profile pictures and found users high in `openness' prefer more aesthetic pictures, while \revchris{users with strong} `agreeableness' and `conscientiousness' 
display more positive emotions in their photos.
Similarly, Cristani et al. (2013) 
studied \revchris{the} relationship between personality of users and the types of pictures they prefer (\eg by examining the types of pictures they like). They found that pictures posted as ``favourite'' by users in Flickr\footnote{\url{https://www.flickr.com/}} were correlated with their personality traits. For example, they found that extrovert individuals show a preference of pictures portraying people while introvert show the opposite preference. 

Other works combine one or more of the above discussed modalities to predict personality traits \cite{kampman2018investigating}.
For example, Kindroglu et al. (2017) 
combine features from audio and visual modalities to predict extraversion and leadership traits.
As of yet, to the best of our knowledge, 
ML techniques have not been used to learn typical linguistic properties that assess personality styles at a specific level of behaviors (\ie anaclitic \vs introjective) for clinical psychology purposes.

Therefore, the aim of our study is twofold, namely to
\begin{enumerate*}[(a),afterlabel=\hspace{-1pt}]
\item ~better understand anaclitic versus introjective primary personality traits through the \textbf{analysis} of natural language; and
\item ~investigate possible \textbf{ML solutions} for automatic personality prediction.
\end{enumerate*}

First, we explore important linguistic characteristics of anaclitic versus introjective patients by performing feature analysis on data from intake interviews (transcripts and audio) of depressed patients.
Notwithstanding the explorative nature of feature analysis, there are certain expectations that can be drawn from a recent line of preliminary descriptive research into the different linguistic styles of anaclitic and introjective patients \cite{davalos2017introjective,valdes2015verbal}. 
Since anaclitic patients are highly concerned with the development of meaningful and satisfying interpersonal relationships (\eg expressed through LIWC categories as `social', `family',  `affiliation'), we expect them to use more pronouns in the second and third person such as `you', `he', `she', etc.
\revchris{Conversely,} we expect introjective patients to frequently use the word  `I', since they are rather preoccupied with achieving a differentiated and consolidated identity (\eg LIWC category  `achieve').
Furthermore, we expect anaclitic patients to use more emotional words (\eg LIWC categories  `affect' and  `feel') and speak with an affective invested intonation, whereas more cognitive and causal words (\eg LIWC categories  `CogProc',  `Insight',  `Cause') are expected in introjective patients' speech.
In a nutshell, the first objective is to show how analyzing verbal behaviors coalesce\revchris{s} into \revchris{an} interpretable and meaningful distinction in personality. 

Second, we investigate whether personality classification into anaclitic or introjective can be automated via ML techniques, using basic and more advanced classifiers with diverse data as input (text, audio, questionnaires, and different combinations of these). The aim hereof is to
\begin{enumerate*}[(a)]
\item assess whether ML allows automatic personality style prediction as good (or better) than via the traditionally used self-report questionnaires that are
\revchris{crafted by experts,}
and
\item which data and selected ML method attains the better personality classification performance.
\end{enumerate*}

The remainder of the paper is organized as follows: \secref{sec:methods} describes how the clinical diagnostic Interview (CDI) was conducted, as well as the data pre-processing, the feature extraction and the machine learning models we used. \secref{sec:results} presents the analytical findings followed by the predictive experimental results. \secref{sec:discussionandconclusion} summarizes our conclusion and discussion.


\section{Materials and Methods}
\label{sec:methods}




\subsection{Participants}
The dataset comprises the intake interviews of 79 patients enrolled in the Ghent Psychotherapy Study (GPS), a randomized controlled trial studying the differential efficacy of cognitive behavioral therapy and short-term psychodynamic psychotherapy for depressed patients with a predominant anaclitic or introjective personality style \cite{meganck2017ghent}. 
All patients are diagnosed with Major Depressive Disorder (MDD), assessed by the Hamilton Rating Scale for Depression \cite{hamilton1967development} and Structured Clinical Interview for DSM-IV-TR \cite{first2002structured}. Exclusion criteria were primary diagnosis of substance abuse/dependence, acute psychotic symptoms or suicidal ideation. The sample consists of 50 anaclitic patients (12 males (24\%), 38 females (76\%); age range 20-60 years (mean = 38.3, standard deviation = 12.3)) and 29 introjective patients (12 males (41.4\%), 17 females (58.6\%); age range 21-59 years (mean = 37.6, standard deviation = 11.0)).  

\subsection{Procedures and Measures}
\label{sec:procedureandmeasures}
Before enrollment in the study, patients filled in a large test battery at home consisting of the \emph{Depressive Experience Questionnaire} (DEQ) \cite{blatt1976experiences} and the \emph{Personal Style Inventory} (PSI) \cite{robins1991sociotropy} to assess personality characteristics; and the \emph{Beck Depression Inventory-II (BDI-II)} \cite{beck1996beck}, the \emph{Depression Anxiety Stress Scale} (DASS) \cite{lovibond1996manual}, the \emph{Symptom Checklist-90} (SCL-90) \cite{derogatis1992scl}, the \emph{Inventory of Interpersonal Problems-32} (IIP-32) \cite{horowitz2002iip}, the \emph{Self-rating Inventory for Posttraumatic Stress Disorder} (SIL) \cite{hovens2000handleiding}, 
the \emph{Outcome Questionnaire-45} (OQ-45) \cite{lambert2004outcome}, 
\emph{Experiences in Close Relationships} (ECR) \cite{fraley2000item} and ten emotions on a \emph{Visual Analogue Scale} (VAS) \cite{van2014you} to assess symptoms and complaints related to depression, anxiety, trauma, interpersonal problems, overall well-being and emotional experiences.

The Clinical Diagnostic Interview (CDI) \cite{westen2006clinical} was administered and audio-taped by a member of the GPS research team upon the intake of the patient. The CDI is a semi-structured interview that questions both current clinical complaints and meaningful lifetime events. The interview guide was built using open questions to evoke rich narratives concerning inter- and intra-personal experiences. The raw audio files of these intake interviews were transcribed using preset standards, and the resulting textual transcripts saved as Word documents.

Personality style was assessed through an intensive prototype matching procedure \cite{werbart2014changes} conducted by three independent and trained GPS researchers (including the researcher who conducted the interview). The procedure consists of each researcher rating the CDI individually on a scale of 1 to 5 for both the anaclitic as well as the introjective personality style using prototype vignettes \cite{werbart2014changes}. In a second step, these independent ratings were discussed to reach consensus. To assign a personality style to a participant, a score of at least 3 for one of the personality styles and a minimum of 2 points difference with the score on the other personality style were required. When no agreement could be reached, the difference in consensus scores between dependent and self-critical personality dimensions was less than 2 points, or there was no score of at least 3 on either dimension, patients were excluded from the trial. 

\subsection{Data Preprocessing}
Our data analysis is based on features obtained from the CDI intake interviews, by processing both audio files and human produced transcripts thereof, as sketched in \figref{fig:featureextraction}. The transcripts (\ie Word documents) of the CDI were segmented into questions, answers, and background noises according to the text formatting. Anonymization was performed by first extracting capitalized words, thus obtaining a list that was manually filtered to retain a list of unique person and location names to anonymize (i.e., replace them with a special \textit{$<$name$>$} or \textit{$<$location$>$} tag). Using the \emph{Natural Language Toolkit} (NLTK) \cite{perkins2010python}, the text was then further tokenized and standardized (e.g., lowercasing, mapping dialect words and numbers to standard language, applying tags for sounds and anonymizations) and eventually appended into a JSON-file. The audio (\ie recorded in MP3-format) of the CDI, was converted to a mono-recording and stored as a WAV-file with a normalized volume, a sample rate of 16 kHz and a bit-depth of 16 bits per sample. Speaker segmentation and speaker clustering \cite{desplanques2017adaptive} was used to separate speech into interviewer, interviewee utterances and non-speech, using the Open Speech and Music Interpretation by Large Space Extraction (openSMILE).\footnote{\url{https://www.audeering.com/research/opensmile/}} openSMILE is a modular and flexible toolkit used for feature extraction from audio files. The resulting segments were stored as separate TextGrid files.

\subsection{Feature extraction and Machine Learning Models}
\label{sec:featurepreprocessingandmlmodels}
\begin{figure}[h!]
\includegraphics[width=\linewidth]{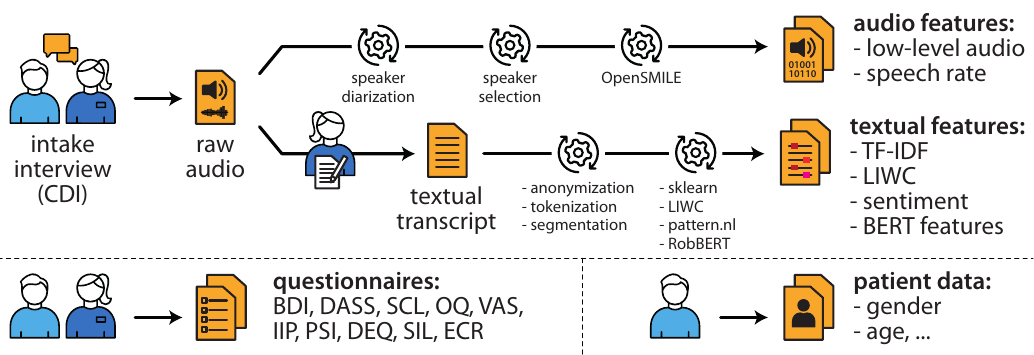}
\caption{The input data and the feature extraction pipeline. The features extracted from the CDI intake interviews include: TF-IDF, LIWC, BERT features, and audio features. The questionnaire features are filled by the patients.}
\label{fig:featureextraction}
\end{figure}

In this section, we discuss how the features extracted from the CDI intake interviews are further processed before they can be fed to machine learning models. Additionally, we discuss which machine learning techniques were used in order to create models that classify the personality styles of patients. \Figref{fig:featureextraction} shows the complete feature extraction pipeline. We built our machine learning models based on the feature categories we will discuss next. 

\subsubsection{Questionnaire features}
As briefly introduced in \secref{sec:procedureandmeasures}, patients filled a large battery of questionnaires before they were enrolled in the study. The joint outputs of all collected questionnaires yielded a feature vector with a total of 358 elements. This feature vector contains both categorical and numerical values. It is then fed to several classifiers for determining the personality trait of the the patients. For our questionnaire feature set, we trained a simple logistic regression (LR), a random forest (RF), and a CatBoost (CB) classifier. We use the LR as our baseline because it is one of the most basic and popular algorithms to solve binary classification problems. The second ML model, random forest (RF) is an ensemble of decision tree classifiers, in which it is ensured that each tree is an acceptably good classifier on its own and that the trees are uncorrelated. Finally, we use CatBoost as it is specifically designed for dealing with categorical variables and has proven to attain state-of-the-art results for classification tasks based on tabular data.

\subsubsection{TF-IDF features}
To process the textual transcripts, for each word in a document we calculated the TF-IDF scores as features \cite{ramos2003using}. An inverse document frequency (IDF) weight indicates the relevance of a certain word within a certain document, with lower weights being associated with words that are common in the corpus and thus have small discriminating power. This word IDF is then multiplied by the term frequency (TF) in a particular document to obtain the TF-IDF score of that word for the respective document. Calculating those TF-IDF scores for each word, we thus summarize a document as a vector of these per-word scores, indicating which words are important in the document and which are not. This procedure was applied to the interview transcriptions, converting each document to a high-dimensional sparse real-valued vector, with a value for each different word encountered. The same TF-IDF scores were also calculated for N-grams. An N-gram is a sequence of N consecutive words appearing in a document, thus allowing longer grammatical constructions to be captured as well. In sum, the set of features for the TF-IDF based classifiers consists of a vector for each document containing the TF-IDF score for that document of each N-gram (N = 1, 2, 3) in the corpus (\ie 840,834 elements).

We used logistic regression (LR), random forest (RF) and CatBoost (CB) to classify the patients. Rather than considering all transcribed speech of the patient in the complete interview as a whole, we also experimented by considering “chunking” their utterances in the individual answers, treating each per-answer “chunk” as a separate document sample. We then applied majority voting on the per-chunk classification to predict the personality of a given patient. 

\subsubsection{Psychological features}
\label{sec:psychologicalfeatures}
Next, we turned to text processing techniques specifically tailored to an application with features known to be indicative of psychological characteristics and/or sentiment. First, we employed the 2015 Dutch version of the \emph{Linguistic Inquiry Word Count-Lexicon} (LIWC), a computerized text analysis tool for counting words in 73 psychology-relevant word categories, established through psychometric studies on linguistics \cite{tausczik2010psychological}. The output of the LIWC tool is a vector that contains the absolute number of times a word from each category was used. We normalized these absolute count values to fractions of all uttered words in patient's answers. Mann-Whitney-Wilcoxon tests were performed to establish which categories are relevant discriminators between the two personality styles \cite{delacre2017psychologists} where the significance of each category was expressed as a $p$-value. The results will be discussed in \secref{sec:analysis}. 

Second, we used the sentiment analysis functionality from the \emph{pattern.nl} library. The sentiment-function takes a sentence or paragraph as input and returns a tuple of two values, subjectivity and polarity. The subjectivity has a value between 0.0 and 1.0, with low values when the text is objective (\eg “The movie was about squirrels.”) and higher values indicating subjective statements (e.g., “The movie was very good!”). The polarity indicates whether the sentiment expressed in a sentence (\eg “I absolutely hated the movie”, “it was quite good”) is negative or positive, with values ranging between –1.0 and +1.0. The greater the absolute value, the stronger the sentiment. To extract sentiment features from transcripts, we first compute the polarity and subjectivity of each sentence in the transcript, then use the mean and standard deviation for both dimensions as the feature set. Again, Mann-Whitney-Wilcoxon tests were performed to examine whether the differences in terms of subjectivity and polarity between the two personality styles were significant and significance was expressed as a $p$-value. See \secref{sec:analysis} for details. 


\subsubsection{BERT features}

Another feature set we extracted makes use of advanced deep neural models called transformers \cite{vaswani2017attention}, which represent words depending on their semantic role in the context of the text. Bidirectional Encoder Representations from transformers (BERT) \cite{devlin2018bert} is the most popular pre-trained language model that has been widely used to extract rich representations from textual data and proved to attain state-of-the-art results in several  NLP tasks such as sentiment analysis, question answering, machine translation, \etc Moreover, several downstream tasks such as depression detection \cite{rodrigues2019multimodal}, suicide risk assessment \cite{matero2019suicide}, and personality detection \cite{kazameini2020personality} have benefited from such context-aware representations. Since our data is in Dutch, we use RobBERT \cite{delobelle2020robbert}, a variant of BERT that is pretrained on Dutch language, which has achieved state-of-the-art performance for a wide range of tasks in Dutch. Typically, a BERT based model gets an input text (\ie the textual transcripts) prepended by a special token called \emph{CLS}, and produces output representations (\ie a vector of 768 elements) for each of the words in the input and the CLS token. The CLS token is pretrained to represent the entire input sequence. Thus, the vector representation of CLS is used as the extracted feature for the input transcript. Another method is to combine the word representations excluding the CLS token, for example, taking the average or the maximum of over all vector representations. In this paper, we experiment with both CLS representation and max-pooled representation.

Since the input to BERT models is limited to only 512 tokens, we segment our textual transcriptions into chunks of 512 tokens. For determining the class label of a transcript, we look at the predicted labels of its chunks and decide the final class label based on majority voting over the individual of the chunks labels. For this feature set, we built a simple logistic regression model.

\subsubsection{Audio features}
Since in depression studies, noticeable differences in speech production have been suggested as potential biomarkers for depression \cite{mundt2012vocal}, we also collected such features to investigate whether they are useful to detect personality differences. As it has also been established that there are language differences between males and females \cite{newman2008gender}, we will also consider gender as a confounding variable. From the audio, features were extracted from the patients' speech using openSMILE, a modular and flexible toolkit used for feature extraction from audio files \cite{huang2018depression}. We chose to use the extended version of the \emph{Geneva Minimalistic Acoustic Parameter Set} (GeMAPS) as configuration, as it obsoletes feature selection \cite{eyben2015geneva}. This resulted in a feature vector with 88 low-level audio descriptors for each segment in which the patient was speaking. These descriptors include frequency-, energy- and amplitude-related parameters, as well as spectral and temporal parameters. These parameters were normalized to have zero mean and unit variance across all segments belonging to one patient. We also calculated the speech rate for the patient, using two different metrics. The first is the so-called \emph{speaking rate}, calculated as the number of words spoken per second, where the time counted includes the hesitations between the words where no words are pronounced. The second is the \emph{articulation rate}, expressed in characters per second, where silences or hesitations where no speech is produced are excluded from the timekeeping. Finally, we also calculated the fraction of time a patient hesitates while speaking and recorded it as the \emph{hesitation fraction}. The audio feature set thus contains 91 features in total (\ie 88 low-level audio features and 3 speech rate-related features). 
Again for this feature set, we experimented with logistic regression, random forest and CatBoost machine learning models.


\section{Results}
\label{sec:results}
In this section, we present and discuss the analytical and experimental findings. \secref{sec:analysis} provides a thorough analysis of both textual and audio modalities to distinguish anaclitic from introjective personality traits. \secref{sec:mlmodelresults} reports the results of the predictive ML models that were built using the features introduced in \secref{sec:featurepreprocessingandmlmodels}.
\subsection{Analysis}
\label{sec:analysis}
\subsubsection{Text Analysis}

\begin{figure}[t]
\includegraphics[width=\linewidth]{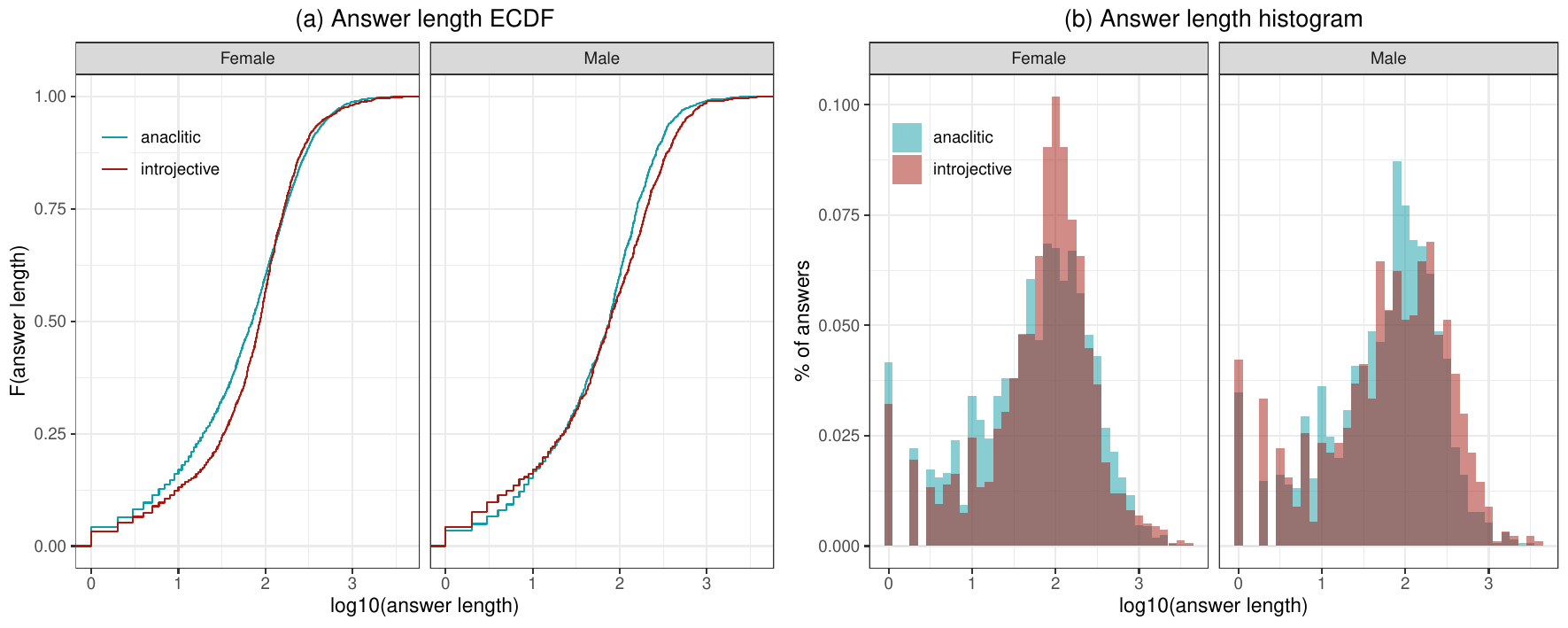}
\caption{Distribution of the answer length (in number of tokens without punctuation) of the patients' aggregated answers in the CDI transcripts: (a) empirical cumulative distribution function (ECDF), (b) histograms}
\label{fig:answerlength}
\end{figure}

\begin{itemize}
\item \textbf{Answer Length} The average transcript length of the CDIs is 15,226 words (disregarding punctuation), with a standard deviation of 5,208 words. The patient's answers constitute the bulk thereof, i.e., 13,556 words (standard deviation of 5,131 words). This standard deviation is fairly large, which can be explained by the rather limited size of the dataset. The distribution of the document length is shown in \figref{fig:answerlength} for the complete transcription set and for the different personality styles separately. \revchris{This figure shows that the length of answers does not seem to significantly differ between personality styles or sexes.}

\begin{figure}[ht]
\centering

\begin{subfigure}{\textwidth}
  \centering
  \includegraphics[width=0.9\linewidth]{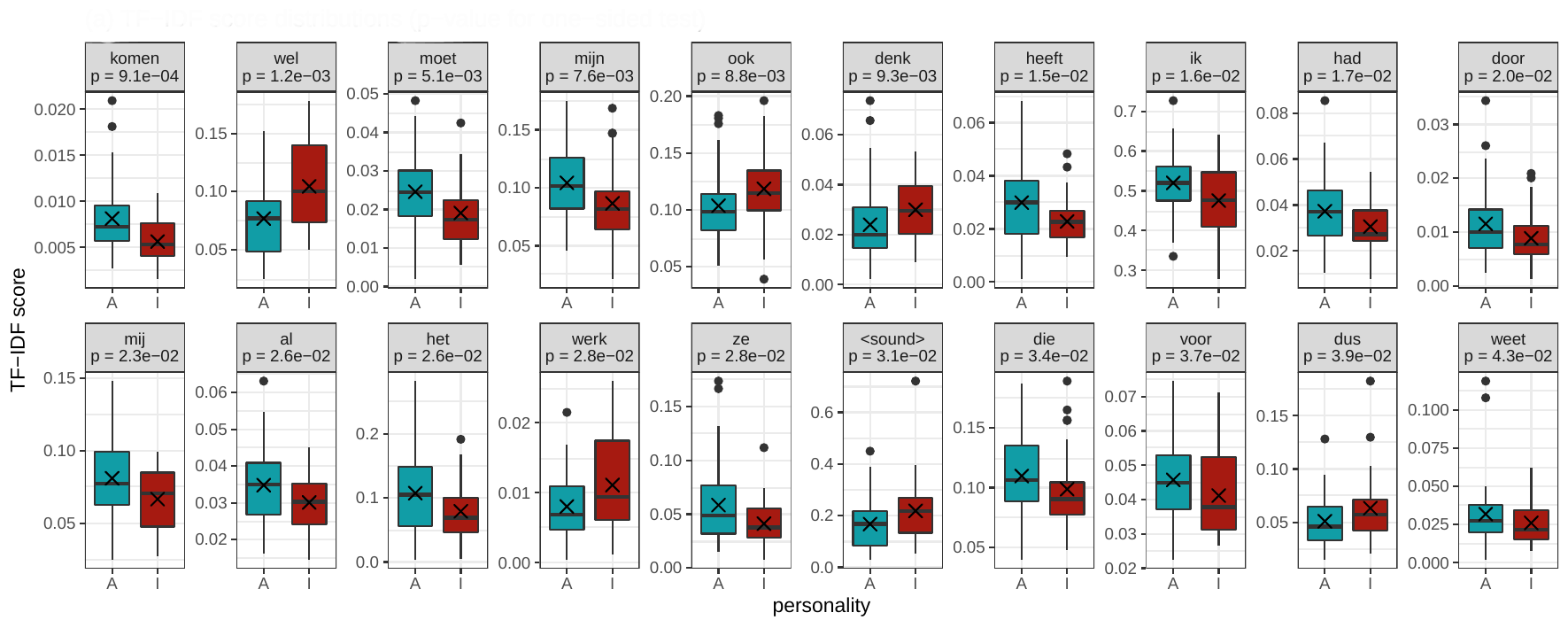}
  \caption{TF-IDF score distributions}
  \label{fig:tfidfanalysis}
\end{subfigure}

\begin{subfigure}{\textwidth}
  \centering
  \includegraphics[width=0.9\linewidth]{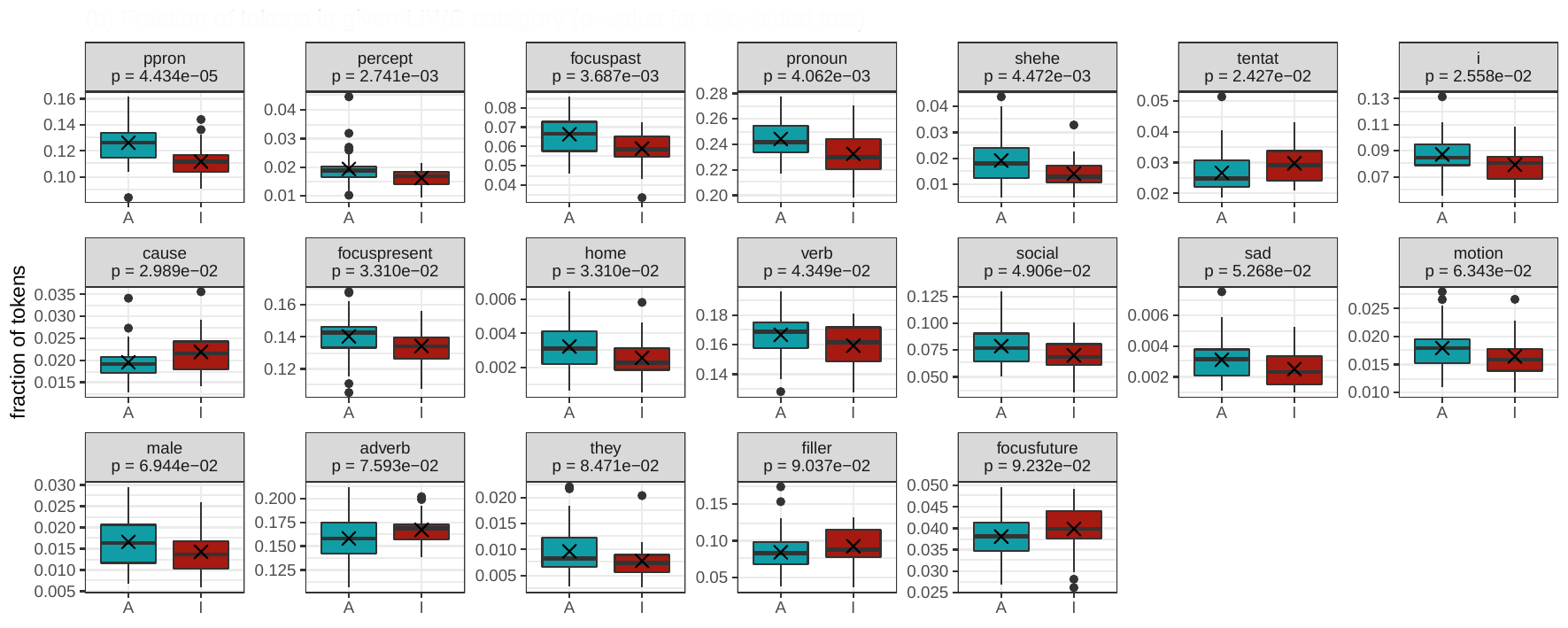}
  \caption{Fraction of tokens in given LIWC category}
  \label{fig:liwcanalysis}
\end{subfigure}

\caption{Lexical features that are potentially useful for distinguishing anaclitic (A) versus inrojective (I) personality styles: (a) TF-IDF score distribution for the most discriminating words, (b) LIWC category frequency (measured as fraction of text tokens). The stated $p$-values result from a Mann-Whitney-Wilcoxon with null hypothesis that the feature distributions are the same for anaclitics and introjectives. Note:  × represents the mean value.}
\label{fig:textualanalysis}
\end{figure}



    \item \textbf{TF-IDF Analysis}: Before considering specific word categories in LIWC, we analyzed the pure vocabulary usage by means of individual words' TF-IDF scores. In \figref{fig:tfidfanalysis}, we plot the distributions of TF-IDF scores in the anaclitic and introjective patient populations for words that (i) occur in all transcripts; and (ii) occur significantly 
    in one of the personality categories, according to a Mann-Whitney-Wilcoxon test at $p < 0.05$. Even though these features consider documents as bags of words, and thus ignore word order or the context in which they occur, we still observe some meaningful insights. We note the higher frequency of personal pronouns — which are notable features in distinguishing personality styles \cite{pennebaker2011secret} 
    — in anaclitics (see “ik (I)”, “mij (me)”, “mijn (my)”, “ze (she/they)”). For introjectives, we further note more prevalent “denk (think)”, “wel (well)”, “ook (also)” which one would expect to occur more in statements reflecting cognitive processes, nuanced reasonings, \etc and thus is in accordance with our a priori expectations. We also remark that introjectives utter more non-word sounds (the “\Tag{sound}” marker), vocalized as “uh”, “pff” and the like, which could be seen as hesitations that reflect an ongoing cognitive process while talking. 
    
\begin{table}[h]
\centering
\caption{LIWC categories whose relative frequency among the words used by introjective versus anaclitic patients is significantly different, at least at $p < 0.10$ . }
\label{tab:liwcresults}
\begin{tabular}{cccc}
\toprule
\textbf{Rank}        & \textbf{Category}                   & \textbf{$p$-value}     & \textbf{Dominant personality style} \\
\midrule
1                    & {\color[HTML]{1190A6} ppron}        & 4.43E–05             & {\color[HTML]{1190A6} anaclitic}    \\
2                    & {\color[HTML]{1190A6} percept}      & 2.74E–03             & {\color[HTML]{1190A6} anaclitic}    \\
3                    & {\color[HTML]{1190A6} focuspast}    & 3.69E–03             & {\color[HTML]{1190A6} anaclitic}    \\
4                    & {\color[HTML]{1190A6} pronoun}      & 4.06E–03             & {\color[HTML]{1190A6} anaclitic}    \\
5                    & {\color[HTML]{1190A6} hear}         & 4.19E–03             & {\color[HTML]{1190A6} anaclitic}    \\
6                    & {\color[HTML]{1190A6} shehe}        & 4.47E–03             & {\color[HTML]{1190A6} anaclitic}    \\
7                    & {\color[HTML]{A62711} tentat}       & 2.43E–02             & {\color[HTML]{A62711} introjective} \\
8                    & {\color[HTML]{1190A6} i}            & 2.56E–02             & {\color[HTML]{1190A6} anaclitic}    \\
9                    & {\color[HTML]{A62711} cause}        & 2.99E–02             & {\color[HTML]{A62711} introjective} \\
10                   & {\color[HTML]{1190A6} focuspresent} & 3.31E–02             & {\color[HTML]{1190A6} anaclitic}    \\
11                   & {\color[HTML]{1190A6} home}         & 3.31E–02             & {\color[HTML]{1190A6} anaclitic}    \\
12                   & {\color[HTML]{1190A6} body}         & 4.24E–02             & {\color[HTML]{1190A6} anaclitic}    \\
13                   & {\color[HTML]{1190A6} verb}         & 4.35E–02             & {\color[HTML]{1190A6} anaclitic}    \\
14                   & {\color[HTML]{1190A6} social}       & 4.91E–02             & {\color[HTML]{1190A6} anaclitic}    \\
15                   & {\color[HTML]{1190A6} sad}          & 5.27E–02             & {\color[HTML]{1190A6} anaclitic}    \\
16                   & {\color[HTML]{1190A6} motion}       & 6.34E–02             & {\color[HTML]{1190A6} anaclitic}    \\
17                   & {\color[HTML]{1190A6} male}         & 6.94E–02             & {\color[HTML]{1190A6} anaclitic}    \\
18                   & {\color[HTML]{A62711} adverb}       & 7.59E–02             & {\color[HTML]{A62711} introjective} \\
19                   & {\color[HTML]{1190A6} they}         & 8.47E–02             & {\color[HTML]{1190A6} anaclitic}    \\
20                   & {\color[HTML]{A62711} filler}       & 9.04E–02             & {\color[HTML]{A62711} introjective} \\
21                   & {\color[HTML]{A62711} focusfuture}  & 9.23E–02             & {\color[HTML]{A62711} introjective} \\
22                   & {\color[HTML]{A62711} death}        & 9.63E–02             & {\color[HTML]{A62711} introjective} \\
\bottomrule
\end{tabular}
\end{table}

    \item \textbf{LIWC Analysis}: Because of the variation in interview length (see earlier), instead of absolute word counts, we opted for relative usage of LIWC word categories (i.e., the fraction of words used belonging to a given LIWC category). Analysis of these statistics is summarized in \tabref{tab:liwcresults}, which reports the $p$-values of a Mann-Whitney-Wilcoxon test with null hypothesis that the feature distribution in anaclitic and introjective patient populations is the same. Note that we only consider word categories that are used at least once by all patients, and at least 10 times per interview on average over all patients. We find that personal pronouns ( `ppron',  `shehe',  `I',  `they') as well as pronouns ( `pronoun') in general, comprise a larger fraction of uttered words for anaclitics than for introjectives. Further, also the categories  `home' and  `sad' are significantly more prominent in anaclitic than introjective patients (at $p < 0.05$ ). In line with our expectations, words belonging to the cognitive process sub-categories  `cause' (causation, e.g.,  `because',  `effect') and  `tentat' (tentative, e.g.,  `maybe',  `perhaps') are significantly (at $p < 0.05$) more common in introjective patients' speech compared to anaclitic patients and, vice versa, words belonging to the category  `social' are more frequently used by anaclitic patients. Another interesting finding is that anaclitic patients seem to be preoccupied with the present and past (categories  `focuspresent' and  `focuspast'), while introjective patients seem more invested in the future (category  `focusfuture', only marginally significant at $p < 0.10$). \Figref{fig:liwcanalysis} shows the distributions of these LIWC category usage frequencies in the form of box plots. This confirms that mainly (personal) pronouns, as well as  `sad' and  `focuspast' are discriminative word categories to distinguish anaclitics from introjectives.

\begin{figure}[h!]
\centering
\includegraphics[width=\linewidth]{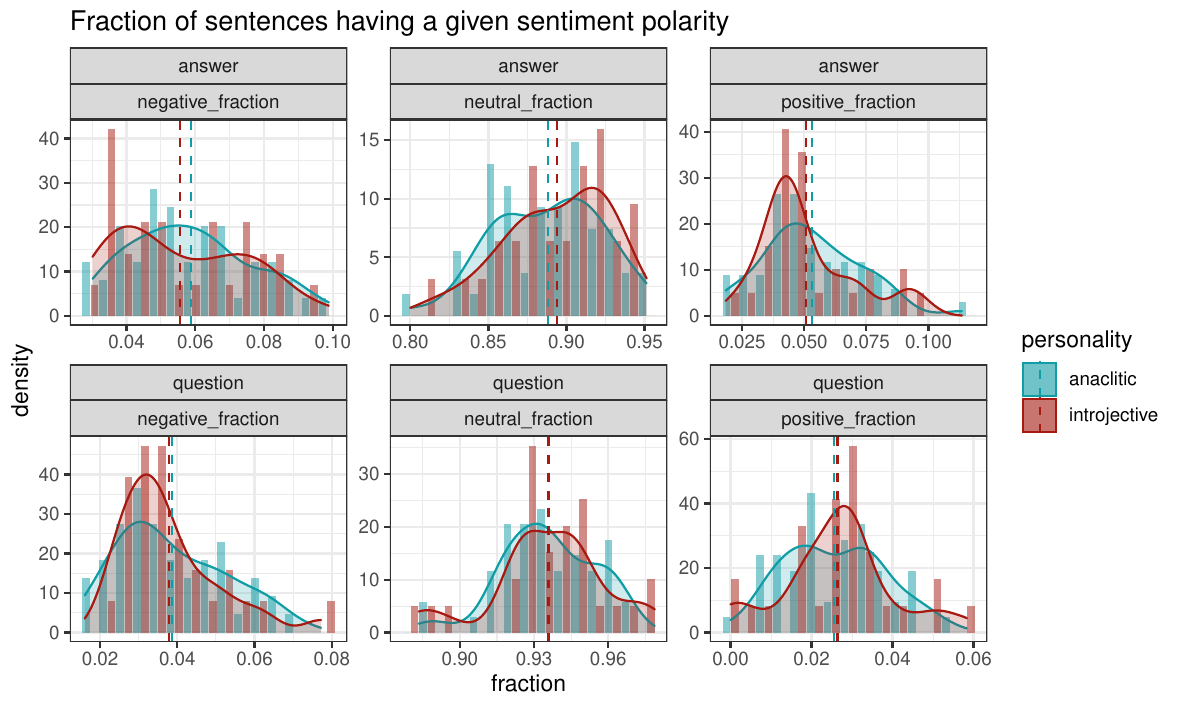}
\caption{Answer and question polarity distributions. Polarity is determined positive for values $> 0.3$, neutral in the range [-0.3, +0.3], and negative if $< –0.3$ 
}
\label{fig:sentimentpolarity}
\end{figure}

    \item \textbf{Sentiment Analysis}: we analyzed the sentiment of the answers given by the patients during their interview with therapists.  First, we discarded short answers \reviewsemere{(\ie containing \textless 10 sentences because the sentiment of a patient is more confidently inferred from longer replies)}. Then, as introduced in \secref{sec:psychologicalfeatures}, we calculated the polarity and subjectivity of patients' answers by taking the average of the polarity and subjectivity of all individual answers. Studying the distributions of the patients' answer polarity and subjectivity did not reveal any statistically significant conclusion.     
    
    Because the answer sentiment analysis did not yield conclusive findings, we opted to use an alternative approach by comparing question-answers pairs \cite{ozkanca2018multi}. From these pairs, we wanted to investigate two \reviewsemere{dimensions: \begin{enumerate*}[(1)]
        \item whether therapists are more likely to ask positive or negative questions to patients with anaclitic or introjective personality styles, and 
        \item whether anaclitic and introjective patients respond more positively or negatively to these questions.
    \end{enumerate*} }
    
    \reviewsemere{To examine the first dimension,} we look at the polarity scores of the questions and answers. Since polarity scores range between $-1$ and $1$, we divided them into three equal parts to reflect positive, neutral and negative polarity. A question/answer was regarded positive if the average polarity over its sentences was larger than $0.3$, negative if it was less than $-0.3$, and neutral otherwise. \Figref{fig:sentimentpolarity} suggests that interviewers/therapists are not biased towards a certain personality style in terms of differentiating the proportions of positive and negative questions.

\end{itemize}
 
 \begin{figure}[h!]
 \centering
\includegraphics[width=0.7\linewidth]{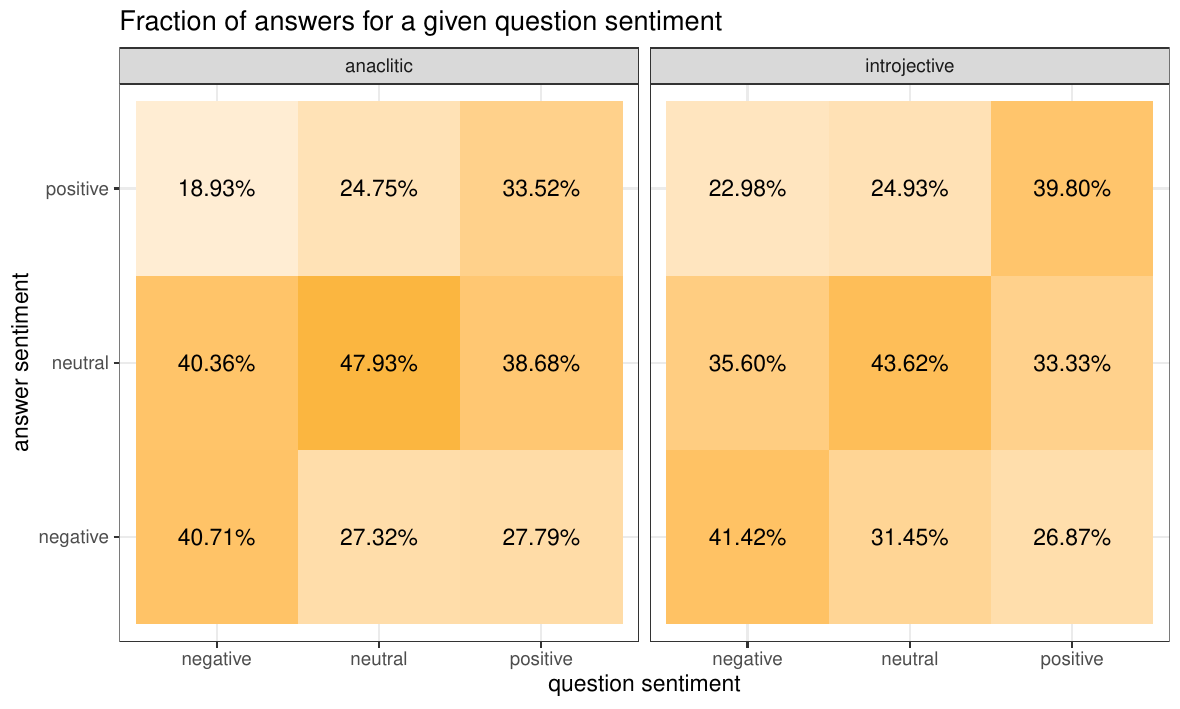}
\caption{Question-answer pair sentiment polarity combinations: per question sentiment, we calculate the fraction of corresponding answers for each of the positive/neutral/negative sentiment categories. The sentiment category of a question/answer is assigned according to the most polarized sentence therein 
}
\label{fig:sentimentheatmap}
\end{figure}

Next, \reviewsemere{to understand the second dimension,} we studied how patients' responses were affected by the polarity of the questions posed to them by the interviewers. \reviewsemere{To achieve this, we first} used the most polarizing sentence in the question/answer as our approximation of its sentiment (negative for an extreme polarity $<-0.3$ , positive for $>$ 0.3, else neutral). Then, for a given question, we calculated the the fraction of answers that fall in each polarity category for both personality styles as depicted in the heat map in \figref{fig:sentimentheatmap}. \reviewsemere{Anaclitic patients seem to stick more to giving neutral answers, while introjectives seem more inclined to follow the question's polarity.}


\subsubsection{Audio Analysis}



We first analyzed the potential difference in audio feature measurements between anaclitic and introjective patients. For the 88 raw audio features, we calculated ANOVA F-scores to determine their respective importance. A high F-score indicates that a feature 
explains a higher fraction of the total variance and thus is more informative. Using this approach, the 10 most relevant features are mean F3-frequency, mean F2-frequency, mean alpha ratio, 20\textsuperscript{th} percentile loudness, mean harmonic difference H1-A3, standard deviation local shimmer (dB), mean local shimmer (dB), mean pitch, mean Hammarberg index (voiced) and mean loudness, respectively. \cite{memon2020acoustic} describes the acoustic meanings of these features. For example, the mean F2 and F3 frequencies indicate a resonant voice while the mean harmonic difference H1-A4 correlates to a creaky voice. However, we did not find indicative features that are meaningfully associated with one of the personality styles, \ie whether any of the features has different group means between introjective versus anaclitic patients. 

\begin{figure}[h!]
\centering
\includegraphics[width=0.8\linewidth]{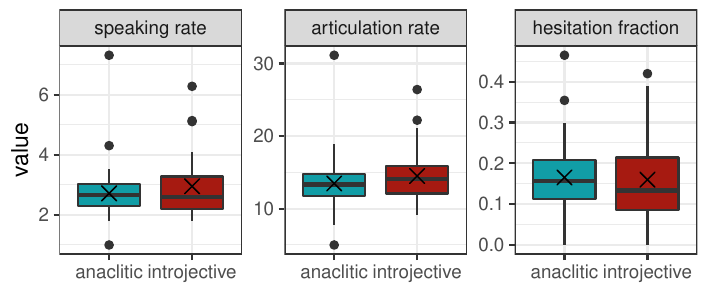}
\caption{High-level audio feature measure distribution in anaclitic and introjective patient populations. Note. × represents the mean value
}
\label{fig:highlevelspeech}
\end{figure}

Additionally, for the more high-level features related to speech rate and hesitations, (a) speaking rate, (b) articulation rate, and (c) hesitation fraction, we performed Mann-Whitney-Wilcoxon tests. Yet, again we found no statistically significant differences in those measures between anaclitic and introjective patient populations, as can also be observed from the largely overlapping box-plots in \figref{fig:highlevelspeech}. 

\subsection{Machine Learning for Automatic Personality Classification}
\label{sec:mlmodelresults}

In this section, we provide the results of the several classification ML models we built to predict personality traits of patients. We use F1-score and Cohen's kappa as evaluation metrics to measure the performance of our models. F1-score can be defined in terms of true/false positives, and true/false negatives. For a given class (i.e., personality style), a true positive (TP) is a positively classified sample (i.e., having the given personality style) that was actually positive and a true negative example (TN) can be defined similarly. A false positive (FP) is a positively classified patient that in reality is negative for the given class, while a false negative is the counterpart of this. As such, the precision (P) of a classifier is the fraction of positive classifications that was actually positive, as shown in equation \equref{eq:1}.  
The recall (R) of a classifier is the fraction of positive training examples that is actually identified as being such (see equation \equref{eq:2}). Finally, the F1-score is calculated as the geometric mean of precision P and recall R as shown in equation \equref{eq:3}.

\begin{equation}
\label{eq:1}
    {P} = \frac{TP}{TP + FP}
\end{equation}
\begin{equation}
\label{eq:2}
    {R} = \frac{TP}{TP + FN}
\end{equation}
\begin{equation}
\label{eq:3}
    {F1} = \frac{2*P*R}{P + R}
\end{equation}

On the other hand, \emph{Kappa} is a measure originally designed as a measure of agreement between two judges, based on the accuracy but corrected for chance agreement. In a classification setting, it is used to evaluate the agreement between the actual and the predicted classes by the model. We use it because it implicitly accounts for class imbalance. Cohen's Kappa is calculated with the following formula: 
\begin{equation}
    \label{eq:kappa}
    \kappa= \frac{P_o - P_e}{1 - P_e}
\end{equation}
where $P_o$ is the overall accuracy of the model (\ie calculated as (($TP + TN)/TP+TN+FN+FP$) and $P_e$ is the expected agreement between the actual classes \& predicted classes. For a binary task like ours, $P_e$ is the sum of $P_{e1}$, the probability of the predictions agreeing with actual values of class 1 (\ie anaclitic) by chance, and $P_{e2}$, the probability of the predictions agreeing with the actual values of class 2 (\ie introjective) by chance. Assuming that the actual class values and model predictions are independent, these probabilities (\ie $P_{e1}$ and $P_{e2}$) are calculated by multiplying the proportion of the actual class and the proportion of the predicted class. $P_e$ is calculated as follows:
\begin{equation}
    \label{eq:pe}
    {P_e}= {P_{e1} + P_{e2}} = {P_{e1, actual} \cdot P_{e1, predicted} + P_{e2, actual} \cdot P_{e2, predicted}}
\end{equation}

To evaluate our proposed classifiers' performance, we used stratified 5-fold cross-validation, where the dataset of 79 patients was split into 5 data segments that (approximately) have the same ratio of anaclitic and introjective patients (63\% resp. 37\%). We used the same splits for each individual classifier, training a classifier model using 4 data segments as training data and evaluating it on the remaining unseen data segment. To report our results, we calculate each metric as an average over the 5-fold cross-validation runs. We repeat this split into 5 folds randomly for 100 times, which results in 100 scores for each metric. Our final results are averages over these iterations and the corresponding standard deviation, as shown in Table 2. It is important to note that the comparison between different models was done using a parallel analysis over the 100 runs. We do a simple t-test between two sets of 100 scores (\ie coming from two models) to check if one model reliably outperforms the other.   

Additionally, since there is a class imbalance in our dataset (\ie 50 anaclitic and 29 introjective patients), we experiment with oversampling techniques to balance the dataset and examine the effect on the results. Particularly, we use synthetic minority oversampling technique (SMOTE) \cite{chawla2002smote} to oversample from the introjective minority class so that the class distribution becomes 1:1. The approach generates new instances by interpolating positive instances (\ie represented by vectors in the feature space) in the minority class that are close to each other. 

The model evaluation for each feature category is shown in \tabref{tab:results}. For the \emph{questionnaire features}, the Logistic regression model achieves the best performance with a macro F1 score of 0.588 and Kappa 0.194. 
Similarly, for the basic \emph{TF-IDF features,} the LR model outperforms the other models with macro F1 score of 0.578 and kappa score of 0.194. The difference in performance between the best questionnaire model and best TF-IDF model is found to be not statistically significant.  Moreover, the LR model based on advanced \emph{BERT features}, which take into account the contextual information, outperforms the best model based on the TF-IDF features. We found that this improvement is statistically significant. The t-test with the null hypothesis that the means of the two models are equal at 0.05 level of significance is strongly rejected ($p<0.001$). Another interesting finding is the max-pooled BERT feature representation yields better classification performance when compared with the the CLS representation confirming conclusions of previous works such as \cite{reimers2019sentence,kim2021self}.           


The best classifier performance using only \emph{one feature category} is achieved using the more advanced features, designed for psychological analysis, \ie LIWC features --- the Random Forest classifier employing these features significantly outperforms all the other classifiers with F1 score of 0.896 and kappa score of 0.801. 
We observe that adding sentiment features and/or patient gender (\ie the psychological features) does not benefit classification performance (\ie F1 score drops to 0.884 and kappa drops to 0.778). However, combining the LIWC features with the questionnaire features produces the best classification performance. The Catboost model that takes the concatenation of the LIWC and questionnaire features as an input achieves 0.93 of F1 score and 0.867 of kappa $\kappa$ score. The improvement is statistically significant with $p=0.001$. \Tabref{tab:t-test-major} in \appref{app:appendex-tables} lists the significance analyses that compare the various models.

Looking at the \emph{audio-based} classification models, we find that the random forest model yields the best results with 0.621 F1 score and 0.268 kappa score. Moreover, this model reliably outperforms the best TF-IDF model. Analyzing our models' means using the t-test results in rejecting the null hypothesis that they are equal with $p<0.001$.

Finally, the oversampling technique (SMOTE) we used helps achieve higher classification performance as can be seen in \tabref{tab:results}. All the best-performing models in each feature category are built using the balanced data. We run a t-test on the best models in each feature category to determine the statistical significance of the stated improvements following dataset balancing. 
The SMOTE models' performance improvement is statistically significant as reported in \tabref{tab:t-test-smote} in \appref{app:appendex-tables}.  

\begin{table}[h!]
\centering
\small
\caption{Classification performance of 5-fold cross-validation results over 100 runs reported as mean $\pm$ standard deviation. The F1 scores are macro averages of the two classes. The best scores for each feature category are \underline{underlined}, and the best scores over all categories are indicated in \textbf{boldface}. }
\label{tab:results}
\begin{tabular}{p{3cm}lllll}
\toprule
Features                        & Models & \multicolumn{2}{c}{imbalanced data}                                            & \multicolumn{2}{c}{balanced data with SMOTE} \\
\cmidrule(lr){3-4} \cmidrule(lr){5-6}

                                &        & \multicolumn{1}{c}{F1}                                     &\multicolumn{1}{c}{Kappa}                               & \multicolumn{1}{c}{F1}                    & \multicolumn{1}{c}{Kappa}               \\ \hline
                               
\multirow{3}{4em}{Questionnaire}  & LR     & 0.586  {\footnotesize±  0.101}           & 0.188 {\footnotesize± 0.194}            & \underline{0.588 {\footnotesize± 0.097}}         & \underline{0.194 {\footnotesize± 0.187}}        \\
                                & RF     & 0.498 {\footnotesize± 0.113}           & 0.115 {\footnotesize± 0.160}            & 0.554 {\footnotesize± 0.118}           & 0.182 {\footnotesize± 0.188}        \\
                                & CB     & 0.453 {\footnotesize± 0.128}           & 0.086 {\footnotesize± 0.171}            & 0.573 {\footnotesize± 0.116}           & 0.194 {\footnotesize± 0.204}       \\
                                \hline

\multirow{3}{4em}{TF-IDF}                          & LR     & 0.570 {\footnotesize± 0.138}           & 0.185 {\footnotesize± 0.228}            & \underline{0.578 {\footnotesize± 0.130}}           & \underline{0.195 {\footnotesize± 0.210}}         \\
                                & RF     & 0.503 {\footnotesize± 0.095}	        & 0.114 {\footnotesize± 0.158}            & 0.524 {\footnotesize± 0.113}	       & 0.128 {\footnotesize± 0.192}         \\
                                & CB     & 0.526 {\footnotesize± 0.112}	        & 0.137 {\footnotesize± 0.188}	        & 0.543 {\footnotesize± 0.121}	       & 0.144 {\footnotesize± 0.209}          \\ 
                                \hline

 \multirow{3}{4em}{LIWC}        & LR     & 0.822 {\footnotesize± 0.129}	        & 0.662 {\footnotesize± 0.229}	        & 0.802 {\footnotesize± 0.129}	       & 0.625 {\footnotesize± 0.230}         \\
                                & RF     & 0.882 {\footnotesize± 0.148}           & 0.779 {\footnotesize± 0.268}            & \underline{0.896 {\footnotesize± 0.127}}           & \underline{0.801 {\footnotesize± 0.238}}       \\
                                 & CB     & 0.838 {\footnotesize± 0.221}	        & 0.717 {\footnotesize± 0.363}	        & 0.873 {\footnotesize± 0.153}	       & 0.763 {\footnotesize± 0.273}      \\
          \hline

\multirow{3}{4em}{Psychological} & LR     & 0.851 {\footnotesize± 0.091}           & 0.709 {\footnotesize± 0.175}            & 0.845 {\footnotesize± 0.112}           & 0.701 {\footnotesize± 0.20}         \\
                                & RF     & 0.874 {\footnotesize± 0.161}           & 0.766 {\footnotesize± 0.295}            & \underline{0.884 {\footnotesize± 0.142}}           & \underline{0.778 {\footnotesize± 0.271}}        \\
                                 & CB     & 0.824 {\footnotesize± 0.221}	        & 0.689 {\footnotesize± 0.366}	        & 0.868 {\footnotesize± 0.172}	       & 0.758 {\footnotesize± 0.306}      \\
\hline

\multirow{1}{10em}{RobBERT-CLS}                     & LR     & 0.575 {\footnotesize± 0.080}           & 0.182 {\footnotesize± 0.149}            & 0.619 {\footnotesize± 0.090}           & 0.248 {\footnotesize± 0.178}         \\
\multirow{1}{10em}{RobBERT-max}             & LR     & 0.606 {\footnotesize± 0.081}           & 0.257 {\footnotesize± 0.127}            & \underline{0.627 {\footnotesize± 0.110}}           & \underline{0.282 {\footnotesize± 0.202}}        \\ \hline

\multirow{3}{4em}{Audio}                           & LR     & 0.388 {\footnotesize± 0.006}           & 0.076 {\footnotesize± 0.253}            & 0.270 {\footnotesize± 0.014}           & 0.082 {\footnotesize± 0.196}            \\
                                & RF     & 0.529 {\footnotesize± 0.143}           & 0.169 {\footnotesize± 0.217}            & \underline{0.621 {\footnotesize± 0.129}}           & \underline{0.268 {\footnotesize± 0.248}}         \\
                                & CB     & 0.556 {\footnotesize± 0.149}           & 0.156 {\footnotesize± 0.288}            & 0.577 {\footnotesize± 0.099}           & 0.173 {\footnotesize± 0.190}          \\
                                \hline

\multirow{3}{4em}{LIWC + Questionnaire}            & LR     & 0.572 {\footnotesize± 0.081}              & 0.159 {\footnotesize± 0.151}                & 0.591 {\footnotesize± 0.094}                  & 0.199 {\footnotesize± 0.180}    \\
                                & RF     & 0.822 {\footnotesize± 0.175}               & 0.674 {\footnotesize± 0.300}                & 0.865 {\footnotesize±  0.138}                 & 0.749 {\footnotesize± 0.247}      \\
                                & CB     & 0.812 {\footnotesize± 0.163}              & 0.653 {\footnotesize± 0.286}                & \textbf{\underline{0.930 {\footnotesize± 0.108}}}                  & \textbf{\underline{0.867 {\footnotesize± 0.206}}}           \\
\bottomrule
\end{tabular}
\end{table}

\section{Discussion and Conclusion}
\label{sec:discussionandconclusion}
The \emph{first aim} of this study was to gain a better understanding of anaclitic versus introjective personality traits through the analysis of natural language. We investigated whether using NLP tools to process natural utterances from patients (as recorded in interviews with a therapist) could uncover typical language features for anaclitic and introjective personality styles. We discovered several language characteristics common to one or the other personality style, such as word use, grammar structures, and language polarity. In line with our expectations, we found that anaclitic patients typically use words related to the interpersonal sphere (personal pronouns, ‘social’ category words). In addition, we found that the category ‘home’ was more common for anaclitic patients. This might be linked to the dominance of interpersonal themes, as the boundary between an anaclitic person and the outer world is weaker than for an introjective person. As a result, they tend to talk more about their home situation. For example: \emph{“I moved to the family home and my parents moved to the house next door.”} This example also helps explain why anaclitic patients — contrary to our a priori expectation — use the first person more frequently than introjective patients. In focusing on their relationships with others, they will use more personal pronouns than introjective patients, for talking about others (\emph{‘shehe’}) in relation to themselves (\emph{‘I’}). Interestingly, we also found that anaclitic patients tend to be more preoccupied with the present and the past (see LIWC categories \emph{‘focuspresent’}, \emph{‘focuspast’}), while introjective patients seem to care more about the future (LIWC category \emph{‘focusfuture’}). With a linguistic pun, this corresponds with hysterics’ emphasis on history (hysteria) \cite{verhaeghe2013does}, 
framing their complaints and symptoms in the context of their life story, much more than an obsessional or introjective patient would, the latter being more likely to isolate these different domains. Minimally, this finding suggest that a descriptive approach to personality styles must be supplemented with theory to understand the phenomenology of the different personality styles and comprehend the clinical usefulness of discriminating between them.

\reviewsemere{From the sentiment analysis, we note that the polarity of the questions asked by the interviewer is not differently distributed between the two personality styles. This suggests that polarized sentiment is not evoked by a different attitude of the therapist based on the patient’s personality. We also note that introjective patients seem to have answers that are more aligned in sentiment with the interviewer, compared to anaclitic subjects who stick to giving more neutral answers (\ie the fraction of answers corresponding in sentiment to that of the question asked is slightly higher for introjectives). Our findings within this regard are preliminary at best, but nonetheless suggest potential avenues for further research into turn taking in therapy.}

The \emph{second goal} of this research was to see whethear personality classification into anaclitic or introjective personality styles can be automated well using machine learning techniques. We conclude from our findings that the advanced psychological features based on LIWC provide the most discriminating features. The best model, which achieves the maximum classification performance, is a random forest utilizing these features. 
When comparing these results to the ones obtained using only questionnaire features, the improvement is significant (increasing from 0.588 to 0.896 and 0.194 to 0.801 for F1 and kappa scores, respectively). \revchris{More importantly, we find that combining these two feature categories further improves the classification performance. A CatBoost model trained on this combined feature set achieves 0.93 F1 score and 0.867 kappa score.} See \tabref{tab:t-test-major} in \appref{app:appendex-tables} for the t-test significance tests.

Classification of personality styles based on of self-report surveys have been criticized in the past \cite{coyne2004promissory,desmet2007histerical} due to concerns about the validity of the questionnaires introduced in \secref{sec:procedureandmeasures}. 
The findings of our study confirm that the questionnaires' discriminatory power is questionable, as demonstrated by the poor performance of machine learning models built with these as the only features when compared to (textual) data extracted directly from patient interviews. \revchris{However, it is worth noting that questionnaires complement advanced textual characteristics, as seen by the rise in F1 score and Kappa score (i.e., LIWC-based model vs. LIWC+Questionnaire model) of an additional 3 and 6 percentage points, respectively.} 
Our results suggest that we should go back to the basics when addressing the assessment issue, meaning that patients’ narratives, what people say when they first come to consult a health care professional, is most indicative for the personality style of the patient. Our results can aid researchers and clinicians in their attempt to determine the personality style of their patients. Despite the good performance of our proposed automatic personality classification, we believe the classifying features that stem from our study should only be used to assist a human psychologist to confirm or challenge his/her thoughts. They should not replace the psychologist as a decision \revchris{taker}.

A limitation of this study is the relative small sample size (N = 79 patients). 
Even though that the main goals of this study were achieved, a larger sample size might have led to more robust and significant findings. Further improvements can thus possibly be established using larger samples. Further, the classifiers in this study were based on only a single intake interview per patient. Thus, we cannot confidently claim that we can 
generalize to intake conversations with structures different from the adopted CDI. Further research should aim to replicate our findings using alternative interviews or natural therapy intake sessions. As we focused specifically on a depressed population, further research could also aim to investigate a more heterogeneous (clinical) sample. Moreover, it would be interesting to investigate associations between personality styles and the process of therapy from diverse therapy orientations in an attempt to discover patterns that have remained obscure in traditional outcome-process research.

\bibliographystyle{clin} 
\bibliography{bibliography}

\section{Appendices}

\appendix
\section{Statistical Significance test details}
\label{app:appendex-tables}

This section provides the statistical significance tests we run to check whether one model reliably outperforms another. \Tabref{tab:t-test-major} shows the $p$-value after running a t-test to compare the best models in each feature category. First, we run each model 100 times, which thus yields 100 kappa scores. We then use a t-test to test the null hypothesis that the means of the two models are equal at a 0.05 level of significance. The only comparison where we found no significant difference is between the questionnaire and TF-IDF-based models with $p$=0.847. We found that the difference was significant for each of the other comparisons.

\begin{table}[h]
\centering
\footnotesize
\caption{The comparison of best-performing models (in terms of kappa score k) from each category using the t-test. Each model is run 100 times yielding 100 kappa scores. We then use a t-test to compare the different models.  Null hypothesis: the means of the two models are equal.} 
\label{tab:t-test-major}

\begin{tabular}{p{2.5cm}p{1cm}p{1cm}p{1cm}p{1cm}p{1cm}p{1cm}p{1cm}p{1.5cm}}
\toprule
                                & Question-naire     & TF-IDF        & LIWC                  & Psycho-logical                & BERT-CLS              & BERT-max              & Audio & LIWC+ Questionnaire\\
                                & $\kappa$=0.194           & $\kappa$=0.194       & $\kappa$=0.801               & $\kappa$=0.778               & $\kappa$=0.248               & $\kappa$=0.282               & $\kappa$=0.268  &$\kappa$=0.867\\ \hline
Questionnaire ($\kappa$=0.194)         &  -                & $p$=0.847     &$p<$0.001      &$p<$0.001      &$p<$0.001      &$p<$0.001      &$p<$0.001   &$p<$0.001    \\
TF-IDF ($\kappa$=0.194)                &  -                & -             &$p<$0.001      &$p<$0.001      &$p<$0.001      &$p<$0.001      &$p<$0.001    &$p<$0.001 \\
LIWC ($\kappa$=0.801)                  &  -                & -             &-                      &$p<$0.001      &$p<$0.001      &$p<$0.001      &$p<$0.001   &$p<$0.001     \\
Psychological ($\kappa$=0.778)         &  -                & -             &-                      &-                      &$p<$0.001      &$p<$0.001      &$p<$0.001   &$p<$0.001     \\
BERT-CLS ($\kappa$=0.248)              &  -                & -             &-                      &-                      &-                      &$p<$0.001      &$p<$0.001   &$p<$0.001     \\
BERT-max ($\kappa$=0.282)              &  -                & -             &-                      &-                      &-                      &-                      &$p<$0.001   &$p<$0.001     \\
Audio ($\kappa$=0.268)                 &  -                & -             &-                      &-                      &-                      &-                      &-   &$p<$0.001    \\ 
\bottomrule
\end{tabular}

\end{table}

Similarly, \tabref{tab:t-test-smote} shows the $p$-values for statistical significance tests using the t-test. The aim is to examine whether the performance increase gained by balancing the dataset is significant. Again, we test the null hypothesis that the means (in terms of kappa scores) of balanced- \vs unbalanced-based models are equal at a 0.05 level of significance. As shown in the table, the null hypothesis was rejected for all the comparisons, indicating that improvements of using SMOTE are significant.

\begin{table}[h]
\centering
\caption{The statistical significance test results for comparing models based on a balanced \vs unbalanced dataset. \revchris{For each feature category, we suse the best performing model type (\ie RF, LR, CB) and compare a model that uses SMOTE (in terms of kappa score) with the counterparts that uses the unbalanced dataset.} 
Each model is run 100 times yielding 100 kappa scores. We then use a t-test to compare the different models.  Null hypothesis: the means of the SMOTE \vs unbalanced dataset-based models are equal.} 
\label{tab:t-test-smote}

\begin{tabular}{lllll}
\toprule
Features                        & \multicolumn{2}{c} {Dataset} &                               \\
\cmidrule(lr){2-3} 
                       & Unbalanced                 & Balanced (SMOTE)        & t-test  \\
                       & {\small \emph{Kappa $\kappa$}}                & {\small \emph{Kappa $\kappa$}}            \\ \hline
Questionnaire          &  0.188                     & 0.194             &$p<$0.001                 \\
TF-IDF                 &  0.185                     & 0.194             &$p<$0.001                \\
LIWC                   &  0.779                     & 0.801             &$p<$0.001                          \\
Psychological          &  0.766                     & 0.778             &$p<$0.001                                      \\
BERT-CLS               &  0.182                     & 0.248             &$p<$0.001                               \\
BERT-max               &  0.257                     & 0.282             &$p<$0.001                                         \\
Audio                  &  0.169                     & 0.268             &$p<$0.001                                       \\ 
LIWC+Ques'ire          &  0.653                     & 0.867             &$p<$0.001       \\ 
\bottomrule
\end{tabular}

\end{table}

\end{document}